\documentclass[a4paper,11pt]{article}
\usepackage{authblk}

\usepackage[utf8]{inputenc} 
\usepackage[T1]{fontenc}    
\usepackage{lmodern}
\usepackage{hyperref}       
\usepackage{url}            
\usepackage{booktabs}       
\usepackage{amsfonts}       
\usepackage{nicefrac}       
\usepackage{microtype}      

\usepackage{amsmath}
\usepackage{amssymb}
\usepackage{amsthm}
\usepackage{algorithm}
\usepackage{algpseudocode}
\usepackage[pdftex]{graphicx}

\theoremstyle{definition}
\newtheorem{definition}{Definition}

\theoremstyle{plain}
\newtheorem{lemma}{Lemma}
\newtheorem{theorem}{Theorem}

\newcommand{\R}{\mathbb{R}}

\DeclareMathOperator*{\argmin}{argmin}
\renewcommand{\d}{\textnormal{d}}

\newcommand{\E}{\mathbb{E}}

\newcommand{\piinv}{\pi^{-1}}
\newcommand{\im}{\textnormal{Im}}

\newcommand{\x}{\mathbf{s}} 
\newcommand{\s}{\mathbf{x}} 
\newcommand{\y}{\mathbf{r}} 
\renewcommand{\r}{\mathbf{y}} 

\newcommand{\DR}{DR}
\newcommand{\fig}{Figure} 
\newcommand{\tsne}{t-SNE}
\newcommand{\didi}{DiDi}
\newcommand{\tab}{Table}
\newcommand{\sect}{section}
\newcommand{\kNN}{kNN}
\newcommand{\KL}{\textnormal{KL}}
\newcommand{\Qknn}{Q_{\text{kNN}}}
\newcommand{\Qknne}{Q_{\text{kNN-E}}}
\newcommand{\Qd}{Q_{\text{data}}}
\newcommand{\Qnd}{Q_{\neg \text{data}}}

\newcommand{\sfrac}[2]{#1/#2}

\graphicspath{{../../experiments/figs/}}

\title{DeepView: Visualizing Classification Boundaries of Deep Neural Networks as Scatter Plots Using Discriminative Dimensionality Reduction}

\author{Alexander Schulz}
\author{Fabian Hinder}
\author{Barbara Hammer}
\affil{Machine Learning Group, Bielefeld University\\ D-33619 Bielefeld, Germany\\
  \texttt{\{aschulz,fhinder,bhammer\}@techfak.uni-bielefeld.de}}
\date{\footnotesize{Preprint of the publication~\cite{ijcai2020-319}. Proofs are provided in the appendix.}}

\begin{document}

\maketitle

\begin{abstract}
  Machine learning algorithms using deep architectures have been able to implement increasingly powerful and successful models. However, they also become increasingly more complex, more difficult to comprehend and easier to fool. So far, 
  most methods in the literature investigate the decision of the model for a single given input datum.
  In this paper, we propose to visualize a part of the decision function of a deep neural network together with a part of the data set in two dimensions with discriminative dimensionality reduction. This enables us to inspect how different properties of the data are treated by the model, such as outliers, adversaries or poisoned data. Further, the presented approach is complementary to the mentioned interpretation methods from the literature and hence might be even more useful in combination with those. Code is available at \url{https://github.com/LucaHermes/DeepView}.
\end{abstract}

\section{Introduction}
\label{sec:intro}

The increasing relevance of methods of AI in diverse areas such as autonomous driving, algorithmic trading, medical diagnoses, or recommender systems is accompanied
by a high potential of vulnerability of these technologies: their use in every-day life in possibly non-stationary environments violates basic assumptions of learning theory such as samples being i.i.d.\ \cite{DBLP:journals/jmlr/MontielRBA18,DBLP:journals/kais/LosingHW18}; adversarial attacks or poisoning can lead to unpredicted behavior of a single decision or the whole model behavior \cite{DBLP:journals/corr/abs-1802-08195,8418594}; 
and skewed sampling of training data can lead to severely biased or unfair machine learning models if no filtering takes place \cite{8452744}.
In combination with legal requirements such as the European Union's general data protection regulation and right of explanation, these issues have led to a recent boost of explainable AI \cite{DBLP:journals/corr/abs-1902-01876},
including sparse local explanations \cite{DBLP:conf/aaai/Ribeiro0G18}, causal modeling \cite{DBLP:journals/corr/Alvarez-MelisJ17},
counterfactual reasoning \cite{ijcai2018-865},
feature relevance determination, or saliency maps \cite{DBLP:journals/corr/abs-1708-08296}, to name just a few approaches. These methods are accompanied by first approaches aiming to quantify what interpretability by humans means \cite{DBLP:journals/corr/abs-1902-00006}.

Yet, many techniques focus on single decisions rather than displaying parts of the decision boundary and the network's generalization behavior in the input space. Up to now, there has been comparably little effort to build on human's astonishing visual perception abilities and to display the behavior of deep networks in a visual plot, which generalizes an extremely natural and intuitive visualization of classiciation prescriptions: a scatter plot. This is always used in
standard textbooks of pattern recognition to explain a classification prescription, it even made it to the cover image in \cite{Scholkopf:2001:LKS:559923}. Yet a method to compute scatter plots, displaying training data, enriched by the decision boundary and network confidence in the plane, does not yet exist for deep networks in high-dimensional input space. Indeed, high dimensionality constitutes the major obstacle, since there do not exist homeomorphic mappings in between topological spaces of different dimensionality. A key challenge is how to efficiently and effectively determine a regularization and which aspects of the data space need to be displayed. In this contribution, we will build on the rich work of non-linear dimensionality reduction techniques, in particular the recent methodology Uniform Manifold Approximation and Projection (UMAP) \cite{umap}, which is mathematically substantiated by a clear topological motivation,
and we propose a pipeline \emph{DeepView}, enabling to display the decision functions of trained deep networks together with benchmark data. To do so, we introduce two central ingredients:
(i)
 we propose a novel discriminative variant of UMAP, which takes into account the information relevant for a 
priorly trained deep network, and we propose a mathematically sound method to compute this information efficiently.  
(ii) Further, we propose a novel way how UMAP can be enriched to also provide an "inverse" mapping, which
abstracts from information, which is not relevant for the deep network (an exact inverse cannot exist, obviously). 
(iii) We demonstrate the effectiveness of the new visualization pipeline \emph{DeepView} for popular deep learning models and data sets.

\section{Related Work}

While many approaches in explainable AI aim to explain single decisions of a model, only few try to provide a large scale view of a trained model or to visualize its decision boundaries. Notable exceptions here constitute \cite{LapWaeBinMonSamMue19,zahavy16} where a projection of data is interpreted with regard to different strategies for solving the task at hand. Although they depict different groups of data, they do not show decision boundaries of a classifier.
Further relevant related work is \cite{bib:NEPL2015-SchGisHam}, where decision boundaries of classifiers are depicted, but this approach is based on a density estimation in the input space, rendering it infeasible for typical domains of deep networks. 

As concerns discriminative dimensionality reduction (\didi), \cite{vennajmlr} 
demonstrated that \didi\ based on the Fisher metric produces better or at least comparable visualizations in relation to other formalizations. Hence, we focus our discussion on \didi\ implemented with the Fisher metric.

%
%

\section{DeepView: visualizing the decision function of a deep network}
\label{sec:deepview}

In the trivial case of two-dimensional data, the visualization of a trained classification model is a straight-forward and useful thing: we can apply the classifier to each position in an area around our data and encode the predicted label and certainty in the background of a scatter plot. This illustrates very clearly how the trained model behaves in each area of the data space. 

For high-dimensional data, however, this is not directly possible for several reasons:
 (i) While we can apply dimensionality reduction (\DR) to visualize data, regular \DR\ techniques will try to preserve all the structure in the data and, such, make critical compromises, for instance preserving brightness differences instead of object relevant properties, in the case of images.
 (ii) The classifier is usually trained in a high-dimensional space. Hence, applying it to every position in the data space is not feasible because of an exponentially growing number of positions. 
 (iii) Finally, visualizing a full high-dimensional classifier in two dimensions is not possible because the decision function is high-dimensional and an unguided projection of it (if possible at all) would yield drastic information loss.

\subsection{DeepView: Proposed Visualization Scheme}

\begin{figure}
  \centering
  \includegraphics[width=0.27\linewidth]{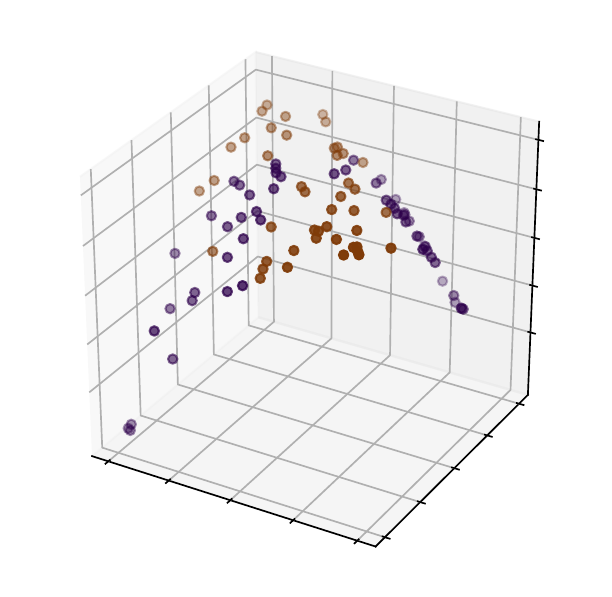}
  \includegraphics[width=0.44\linewidth]{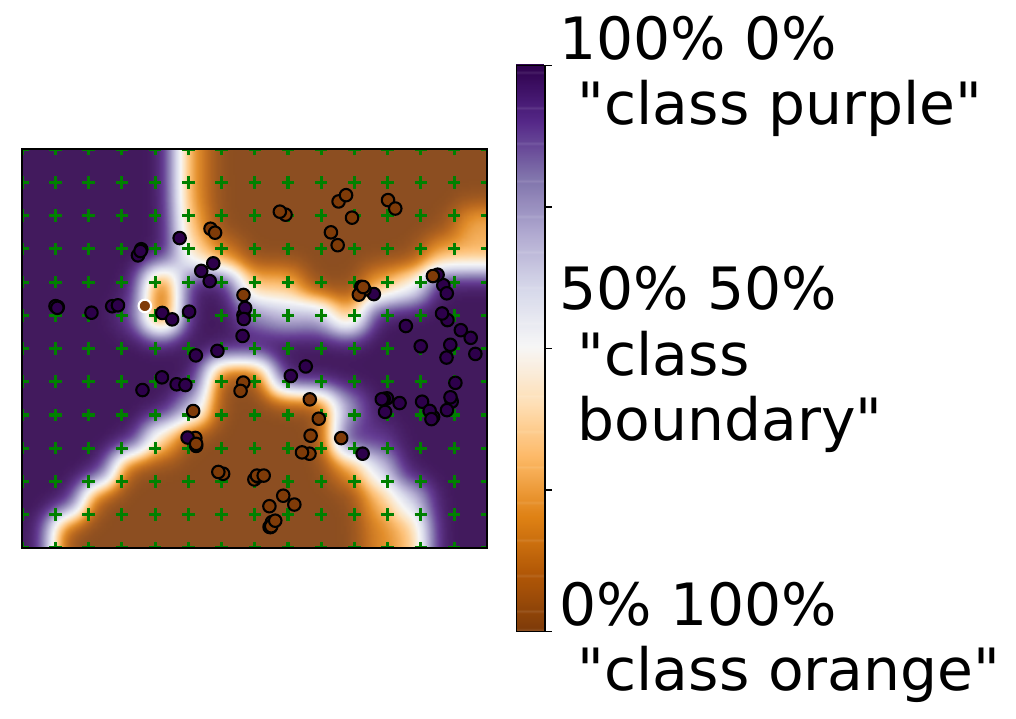}
  \includegraphics[width=0.27\linewidth]{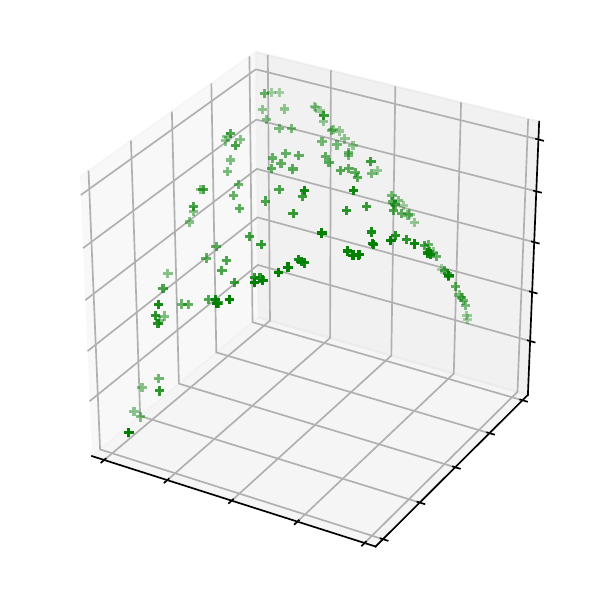}
  \caption{Illustration of the proposed method DeepView. A classifier trained on toy data (left) is visualized with DeepView (middle) utilizing the steps 1-4: data are projected (middle, step 1), a regular grid is created (middle, step 2) and projected with $\piinv$ (right). Then the classifier is applied to the latter (step 3) and the resulting labels and entropy of the predictive distribution are displayed (middle, step 4). The orange island indicates overfitting due to a single orange point.}
  \label{fig:deep_view}
\end{figure}

In this contribution, we develop a scheme that enables us to circumvent these problems and propose feasible approximations for the case of deep neural network classifiers. In particular, we generalize the approach presented in \cite{bib:NEPL2015-SchGisHam} which is restricted to shallow classifiers and (intrinsically) low-dimensional data.
More specifically, we propose to use \didi\ based on the trained classification model (section \ref{sec:didi}), enabling the DR to focus on the aspects of the data which are relevant for the classifier, alleviating problem (i). Further, instead of trying to obtain predictions for the full data space, we develop a scheme to obtain predictions only for a relevant subspace which we then use to visualize the decision function in two dimensions, such, solving problem (ii) and tackling problem (iii) by a reasonable approximation. This is based on inverse dimensionality reduction and is modeled in such a way that it matches the way the DR is constructed (section \ref{sec:invdr}).

We propose to apply the following steps:
\begin{enumerate}
 \item Apply the \didi\ technique Fisher UMAP (developed in \sect\ \ref{sec:didi}) which is based on the underlying deep network to project a data set consisting of points $\s_i$ to two dimensions, yielding $\r_i = \pi(\s_i)$.
 \item Create a tight regular grid of samples $\y_i$ in the two-dimensional space and map it to the high-dimensional space using the approach presented in \sect\ \ref{sec:invdr}, yielding points $\x_i = \piinv(\y_i)$. 
 \item Apply the neural network $f$ to $\x_i$ in order to obtain predictions and certainties. 
 \item Visualize the label together with the entropy of the certainty for each position $\y_i$ in the background of the projection space in order to obtain an approximation of the decision function.
\end{enumerate}

These steps are demonstrated on a toy example in \fig\ \ref{fig:deep_view}.
We propose novel implementations of steps 1 and 2, enabling the resulting approach to visualize deep neural networks.

\section{Dimensionality reduction}
\label{sec:dr}
{
	
	Dimensionality reduction techniques for visualization aim to find mappings ${\pi : (S,d_S) \to \R^d}, d=2,3$, where $(S,d_S)$ is some metric space, such that $\pi$ preserves the information encoded in a set of data points $\s_1,...,\s_n \in S$ as good as possible. The main aspect of a \DR\ method is therefore to find a measure, and hence a cost function, to compare the information contained in two sets of points, allowing us to find a set of points $\r_1,...,\r_n \in \R^d, d=2,3$ encoding approximately the same information. 
While the state of the art approach for performing nonlinear \DR\ is t-Distributed Stochastic Neighbor Embedding (\tsne) \cite{tsne}, recently a novel technique has been developed called UMAP \cite{umap} which produces at least comparable results to \tsne\ and is formulated in a way that we make use of to develop our inverse projection $\piinv$. Hence, we focus on UMAP and introduce some formal details in the following.

UMAP assumes that data is distributed according to a uniform distribution on a Riemannian manifold which may be approximated by a simplicial complex. 
	To find the low dimensional embedding, the problem is restricted to the complex's 1-skeleton, i.e. the probability that two points are connected by an edge. It is assumed that the probability of an edge is induced by the distance between the end points and the local density. In the embedding space ($\R^d,d=2,3$) this is modeled by a student-t-distribution $w_{ij} = (1+a\Vert \r_i - \r_j \Vert^{2b})^{-1}$, where $a$ and $b$ are hyper parameters, in the original space $(S,d_S)$ one uses $v_{ij} = v_{i|j} \perp v_{j|i}$, where $x \perp y = x+y-xy$ is the sum T-conorm and $v_{i|j} = \exp\left(-\max(d(\s_i,\s_j)^2-\rho_i,0)/\sigma_i\right)$, with $\sigma_i$ the $k$-perplexity at $\s_i$ and $\rho_i = \min_{i \neq j} d(\s_i,\s_j)^2$ the distance to the nearest neighbor. 
	
	One then finds $\r_1,...,\r_n \in \R^d, d = 2,3$ by minimizing the Kullback-Leibler divergence of the embedding given the data:
	\begin{align*}
	\argmin_{ \r_1,...,\r_n } \sum_{i \neq j} D_\KL(v_{ij} || w_{ij}) 
	\end{align*}
}

\subsection{Discriminative dimensionality reduction}
\label{sec:didi}
{
	\newcommand{\JS}{\textnormal{JS}}
	\newcommand{\Borel}{\mathfrak{B}}
	\newcommand{\C}{\mathcal{C}}
	\newcommand{\Prob}{\mathbf{P}}
	\renewcommand{\i}{\textbf{i}}

	
	\DR\ usually only takes the observed data points into account. However, since we would like to visualize a classifier $f$ together with data points, it is reasonable to incorporate $f$ as well; such methods are referred to as \didi\ methods. These particularly focus on the structure important for a given classification task \cite{vennajmlr,goldberger,bib:Neurocomputing2015-GisSchHam} and such specify which structure in the data is more important than other. A very promising approach for this purpose is to integrate class information in the distance computation using the Fisher metric and consequently apply a regular \DR\ method on top. Thereby, the Fisher metric is not defined in the parameter space of a model as in \cite{amari1985differential,amari2007methods}, but in the data space \cite{sami,sami2,lisboa}. For this case, \cite{vennajmlr} 
	demonstrated suitability of \didi\ mappings based on the Fisher metric 
	and \cite{bib:NEPL2015-SchGisHam} 
	illustrated its benefit for visualization of shallow classifiers.
	
	Here, we augment the literature by proposing (i) a new derivation of the Fisher metric in the context of \didi, (ii) a new approach to estimate the Fisher metric without needing to compute gradients and (iii) to use $f$ for the probabilistic model instead of a new non-parametric estimate. Together, these enable us to compute a \didi\ visualization for high-dimensional data and deep networks.
	
	
	We will now construct a metric $d$, suitable for our considerations. As it will turn out this is a generalization of the Fisher metric. A significant benefit of our approach is that it is capable of handling non-differentiable, even non-continuous, classifiers and allows a natural approximation that yields a much faster computation, in particular for complicated $f$.
	
	Let $S$ be our source space equipped with a metric $d_S$, $\C$ be the collection of all class labels and $f : S \to \Prob(\C)$ be our classifier, where $\Prob(\C)$ denotes the space of all possible class probabilities. 
	\begin{figure}
		\centering
		\includegraphics[width=0.99\linewidth]{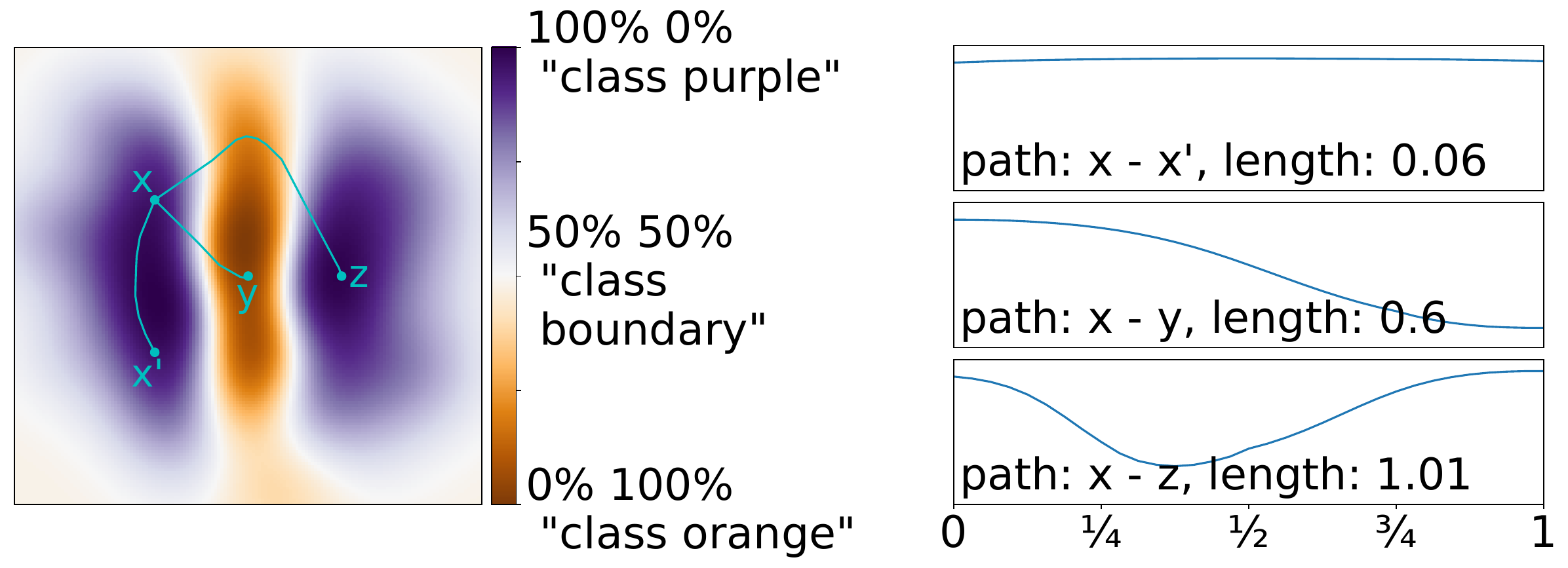}
		\caption{Measuring the distance between points. Left: $f$ and paths. Right: class probability of class purple along the respective paths.}
		\label{fig:illustration_of_d}
	\end{figure}
	In contrast to $d_S$ we would like $d$ to also capture the topological features induced by the decision boundaries of $f$: A first, natural step is to consider $f^* d_{\JS} (s, t) := d_{\JS}(f(s),f(t))$ the so called pullback-metric along $f$, here $d_{\JS}(p,q) = \sqrt{D_\JS(p\Vert q)} := \sqrt{(D_\KL(p \Vert m)+D_\KL(q \Vert m))/2}, m = (p+q)/2$ denotes the Jensen-Shannon-metric, a metric on $\Prob(\C)$. This pullback is an indicator for the boundary features in the sense that it tells us whenever two points are mapped to the same class or not, however it has two issues: (a) it is no proper metric on $S$, i.e. all points with the same assignment are collapsed to a single point, (b) it cannot decide whenever two points belong to the same connected component induced by the decision boundaries, e.g. Figure~\ref{fig:illustration_of_d} $x,x'$ vs. $x,z$. 
	
	To overcome (a) we regularize with $d_S$ using parameter $\lambda \in [0,1]$, i.e. we consider ${(1-\lambda) f^*d_\JS(s,t)+\lambda d_S(s,t)}$. 
	
	To overcome (b) notice that $f^*d_\JS(s,t)$ only captures information of $f$ at $s$ and $t$ but not "between them". A natural way to extend such a local metric, which is also widely used for example in Riemannien geometry and hence information geometry, is by using an arc-length metric, i.e. we use the length of the shortest path connecting two points according to the regularized pullback described above as their distance; its behavior illustrated in Figure~\ref{fig:illustration_of_d}.
	
	
	We therefore arrive at the following formalization:
	\begin{definition}
		\label{def:d}
	For a metric space $(S,d_S)$ together with a classifier $f$ define the \didi-metric $d$ with mixture $\lambda \in [0,1]$ as the arc-length metric induced by the the pullback of the Jensen-Shannon-metric along $f$ regularized with $\d_S$, i.e.
	\begin{align*}
	d(x,y) &= \inf_{\gamma : [0,1] \to S \text{ continuous}, \atop \gamma(0) = x, \gamma(1) = y} L(\gamma) \\
	L(\gamma) &= \sup_{0 = t_0 < \cdots < t_n = 1} \sum_{i = 1}^n (1-\lambda) f^* d_{\JS}(\gamma(t_{i-1}), \gamma(t_{i})) \\&\qquad\qquad\qquad\qquad +\lambda d_S (\gamma(t_{i-1}), \gamma(t_{i})).
	\end{align*}
    \end{definition}
    
	\begin{theorem}
		Let $(S,d_S)$ be a metric space and $f$ a smooth classifier on $S$. Denote by $d$ the \didi-metric with mixture $\lambda = 0$ and by $d_\text{Fisher}$ the Fisher metric (as defined in \cite{sami}) induced by $f$. Then it holds
		\begin{align*}
			d_\text{Fisher}(x,y) = \sqrt{8} d(x,y)
		\end{align*}
		for all $x,y \in S$.
		\begin{proof}
			Follows from \cite{Crooks}; details omitted due to space restrictions
		\end{proof}
	\end{theorem}
   	Note that definition~\ref{def:d} does not need gradient computations unlike \cite{sami}. 
   	Further unlike \cite{bib:NEPL2015-SchGisHam} 
   	it no longer requires density estimation of the data space, which is infeasible for typical domains of deep networks. 
	
	To implement this approach, we follow \cite{sami2} and assume that $d(x,y)$ can be approximated by $n$ equidistant points $p_i = \left( 1- \frac{i}{n}\right)x + \frac{i}{n} y$ on a straight line, i.e.
	\begin{align*}
	d(x,y) \approx \sum_{i = 1}^n (1-\lambda) d_{\JS}\left(f\left(  p_{i-1}  \right),f\left( p_i \right)\right)+\lambda d_S\left( p_{i-1},p_i \right).
	\end{align*}
	These approximations are evaluated in \cite{sami2} with the result that they constitute a good compromise between speed and accuracy for the application of \DR.
	
	In a final step, a nonlinear \DR\ technique is used to project the resulting distances $d(x,y)$ for visualization. For this purpose, the papers \cite{vennajmlr,bib:Neurocomputing2015-GisSchHam} have demonstrated that neighbor embedding methods such as NeRV and t-SNE are particularly well suited. Here, we utilize the novel method UMAP \cite{umap} which belongs to this class of methods and has beneficial properties that we will exploit in the following.
	
		
}

\subsection{Similarity coordinate embedding for inverse dimensionality reduction} 
\label{sec:invdr}
{
	\renewcommand{\x}{\mathbf{x}}
	\renewcommand{\y}{\mathbf{y}}
	
    So far we have discussed the task of finding a mapping $\pi : (S,d_S) \to \R^d, d=2,3$. Now we are interested in finding a reverse map $\piinv : \R^d \to (S,d_S)$ that acts as a pseudo-inverse of $\pi$. In particular we would like to obtain $\piinv$ in a natural way, under the assumption that $\pi$ is given by UMAP. 
    
    In some sense $\piinv$ performs an out of sample extension in the opposite direction; therefore let us first consider a "usual" out of sample extension: Suppose we are given some new sample $\x \in S$ and let $v_i(\x)$ represent the probability that $\x_i$ and $\x$ are close or similar (i.e. $v_i(\x_j) = v_{ij}$), then UMAP aims to find $\y \in \R^d$ with $w_i(\y)$ representing the probability that $\y_i$ and $\y$ are close or similar, by minimize the Kullback-Leibler divergence of $v_i(\x)$ and $w_i(\y)$. 
    %
    Following a Bayesian perspective, to determine $\piinv$ we interchange the role of $\x$ and $\y$ and arrive at
    \begin{align*}
    \piinv(\y) &:= \argmin_{\x \in S} \sum_{i = 1}^n D_\KL(w_i(\y)||v_i(\x))
    \end{align*} 
    where we use $v_i(\x) = \exp(-d_S(\theta_i,\x)^2/\sigma_i)$ and $w_i(\y) = (1+a\Vert \rho_i-\y \Vert^{2b})^{-1}$ as in the case of UMAP, where we have to find $\theta_1,...,\theta_n$ and $\rho_1,...,\rho_n$ such that $\piinv$ fits $\pi$ on our observations, i.e. $d_S(\piinv(\pi(\x_i)),\x_i) \to \min$. 
    
    To make our approach feasible we have to find a way to compute $\piinv$ in reasonable time. Since this heavily depends on $d_S$ we may only consider two examples in more detail: $d_S$ is the Euclidean metric or a Riemannian metric.

    Supposing that $S = \R^D$ with $d \ll D$ and $d_S = \Vert \cdot \Vert_2$ is the Euclidean metric. 
    \begin{theorem}
    	Let $\x_1,...,\x_n \in \R^D$ be source points and $\y_1,...,\y_n \in \R^d$ their corresponding projections. Denote by $f(\x,\y) = \sum_{i = 1}^n D_\KL(w_i(\y)||v_i(\x))$ the cost function of $\piinv$ and by $\hat{f}(\x,\y) = \sum_{i = 1}^n w_i(\y) \Vert \theta_i-\x\Vert^2/\sigma_i$. Then it holds $\hat{f}(\x,\y) < f(\x,\y)$. Furthermore under the assumption of Gaussian noise in the input space in mean it holds
    	\[f(\x,\y) - \hat{f}(\x,\y) \in \mathcal{O}(\exp(-D/2)),\] i.e. $\hat{f}(\x,\y)$ converges in mean exponentially fast to $f(\x,\y)$ as the number of dimensions increases.
    	
    	Furthermore it holds
    	\begin{align*}
    		\argmin_{\x \in S} \hat{f}(\x,\y) = \sum_{i = 1}^n \frac{\sfrac{w_i(\y)}{\sigma_i}}{ \sum_{j = 1}^n \sfrac{w_j(\y)}{\sigma_j} } \cdot \theta_i.
    	\end{align*}
    	\begin{proof}
    		All proofs are omitted due to space restrictions 
    	\end{proof}
    \end{theorem}
    Using this theorem we see that approximation $\piinv$ by a radial basis-function network is well suited.

    To generalize this approach to arbitrary Riemannian manifolds recall that those are locally, approximately given by inner product spaces. It is therefore enough to generalize our approach to  arbitrary, finite dimensional, inner product spaces:

    \begin{lemma}
	   Let $S$ be a finite dimensional real vector space. Let $d : S \times S \to \R$ be a metric induced by an inner product and $X$ be a $S$-valued random variable. Then it holds
	   \begin{align*}
	       {\argmin_{\x \in S} \E\left[ d(X,\x)^2 \right] = \argmin_{\x \in S} \E\left[ \Vert X - \x \Vert_2^2 \right]}.
	   \end{align*}
    \end{lemma}

    So if we consider a Riemannian manifold $(M,d_M)$ and we approximate $d_M$ at $\x^*$, the point we are searching for, we obtain the same formula as in the euclidean case. 
    
    Furthermore when training $\piinv$ using Euclidean metric, the result was comparable to the case where we trained with an local approximation of $d_S$ using Fisher-matrices.
}

\section{Experiments}
\label{sec:experiments}

In this section we apply the proposed method DeepView to visualize classifiers trained on the datasets CIFAR-10 and Fashion-MNIST and demonstrate exemplary findings in the presence of adversarial and backdoor attacks.
Before it can be applied, however, it is important to investigate how accurate this visualization is.


\subsection{Evaluating the Proposed Visualization}
\label{sec:eval}

Here, two questions need to be answered: (i) how well does $\pi$ show the view of the classifier on the data and (ii) how well is $\piinv$ depicting the decisions of the classifier in these areas?

Addressing question (i), we pursue the following intuition: If the projection $\pi$ takes the decision function of the classifier properly into account, then the classification decisions should be partially represented in the structure of the projected data points. I.e.\ points that are close to each other should be points that are classified similarity by the classification model.
We can verify the degree to how much this is true by evaluating the accuracy of a simple classifier trained in the projection space using the labels of the deep network. For this purpose, we utilize the leave-one-out error of a k-nearest neighbor (\kNN) classifier with $k=5$ being a standard choice and refer to this measure as $\Qknn$. When we consider UMAP based on the Euclidean metric, we denote this measure as $\Qknne$

(ii) Evaluating $\piinv$ is less straight-forward. Here we suggest a scheme to evaluate the quality of the depicted decision function on the positions of the data points. For every point $\r_i$, compare the classification label of its original counterpart and the classification label of its inverse projection. More formally, we calculate the accordance of $f(\piinv(\r_i))$ and $f(\s_i)$.
Depending on the selected points, such an evaluation will have different implications: Using pairs $(\s_i,\r_i)$ that have been employed for training $\piinv$ will result in an assessment of the quality at the positions of the data points. Using point pairs that have not been utilized for training $\piinv$ will rate the quality of the mapping at positions not seen before, i.e.\ of areas without data. Both are useful since they tell us how accurate the visualization is in areas where data points are available and in those where this is not the case. We will refer with $\Qd$ to the former and with $\Qnd$ to the latter.

We evaluate the resulting visualizations with these scores, where we use $70\%$ of the data for training $\pi$.

\subsection{Hyperparameter Selection}

We choose the amount of Euclidean regularization $\lambda$ in the following way: we evaluate $\pi$ for $\Qknn$ with $\lambda\in[0.2,0.8]$ and choose the largest one that does not degrade $\Qknn$ significantly. As a result we set $\lambda = 0.65$ for both data sets.
As concerns UMAP, we set \textit{n\_neighbors} $ = 30$ in all cases and \textit{min\_dist} to $1$ for Fashion-MNIST.
For $\piinv$, we set $a$ to the smallest value that does not lead to a large drop in $\Qd$ and $b=1$.

\begin{figure}
  \centering
  \includegraphics[width=0.99\linewidth]{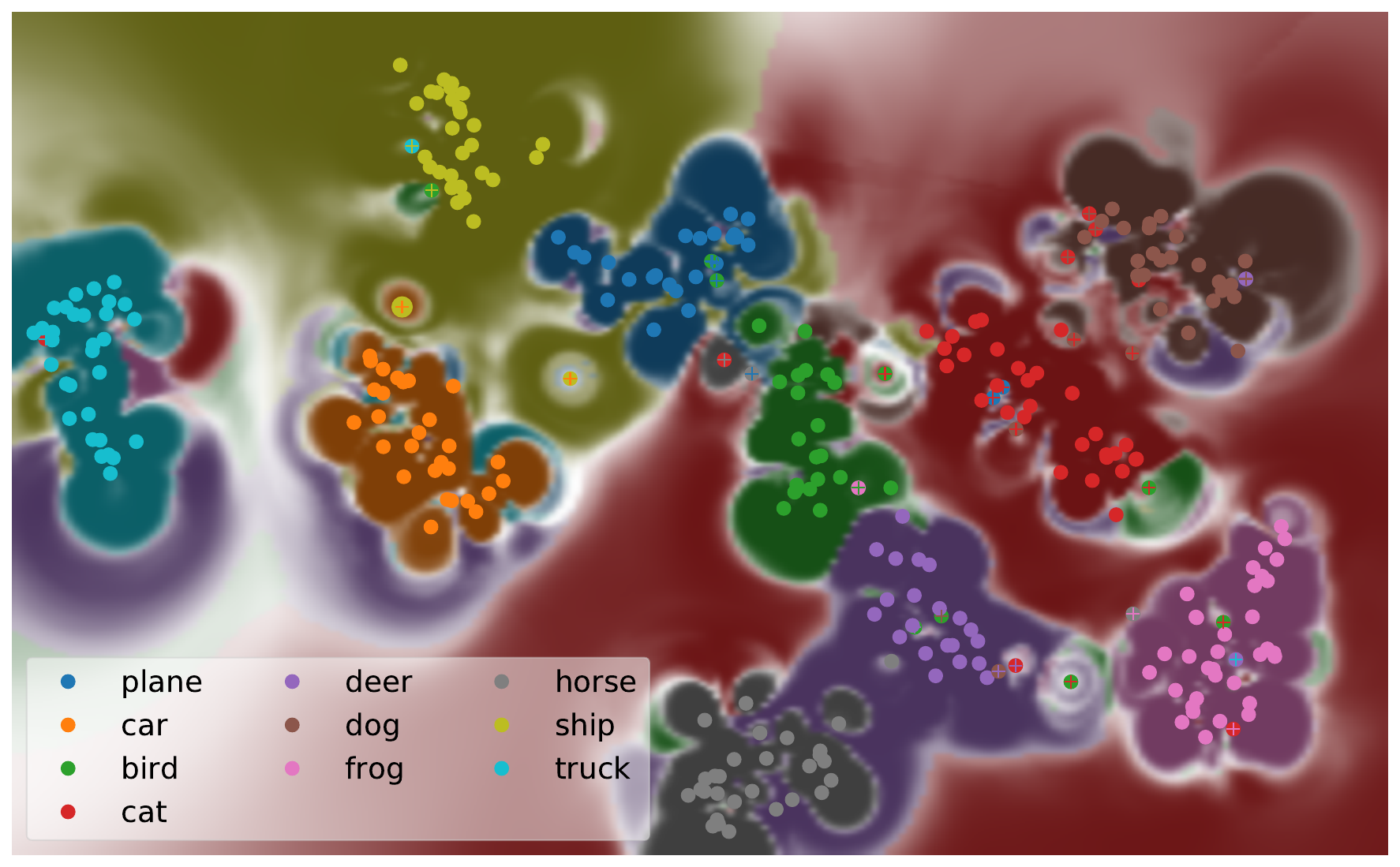}
  \caption{Visualization of a ResNet-20 model trained on the CIFAR-10 data set together with a subset of the test data. The big yellow circle depicts an adversarial example.}
  \label{fig:cifar10_vis}
\end{figure}

\subsection{Visualizing a ResNet-20 Network Trained on the CIFAR-10 Dataset}

The \emph{CIFAR-10 data set} consists of 32x32 color images with 10 classes (see \fig\ \ref{fig:cifar10_vis}). The training set contains 50.000 examples and the present implementation\footnote{We employ the implementation from \url{https://github.com/akamaster/pytorch_resnet_cifar10}.} has an accuracy of $91.7\%$ on the 10.000 test images, using a pre-trained residual network (ResNet) with 20 layers \cite{resnets}.

The result of DeepView applied to a subset of 300 points selected randomly from the test set and the ResNet-20 network is shown in \fig\ \ref{fig:cifar10_vis}. Each point corresponds to one image and the color indicates its original label. If classified differently by the deep net, a cross in the according color indicates that label. The color of the background depicts the classification of the net in that area and the intensity encodes the certainty.

First of all, we consider the evaluation of the resulting visualization and, second, discuss the information
than can be drawn from it.
The former is summarized in \tab\ \ref{tab:eval_res}. For the sake of completeness, we also evaluate the UMAP projection based on the Euclidean metric  ($\Qknne$). The accuracy in this case amounts to $18.3\%$, which makes clear that such an embedding is not useful in this case. With the Fisher metric, however, the accuracy is $96.3\%$ indicating that the projection space very well resembles the classification behavior of our model.
Concerning the visualization of the decision function, the close to perfect $\Qd$ demonstrates that the visualization is very accurate at the given data points. For the vicinity of the data points, $\Qnd$ asses an accuracy $83.3\%$.

\begin{figure}
  \centering
  \includegraphics[width=0.3\linewidth, trim=0 10 0 0]{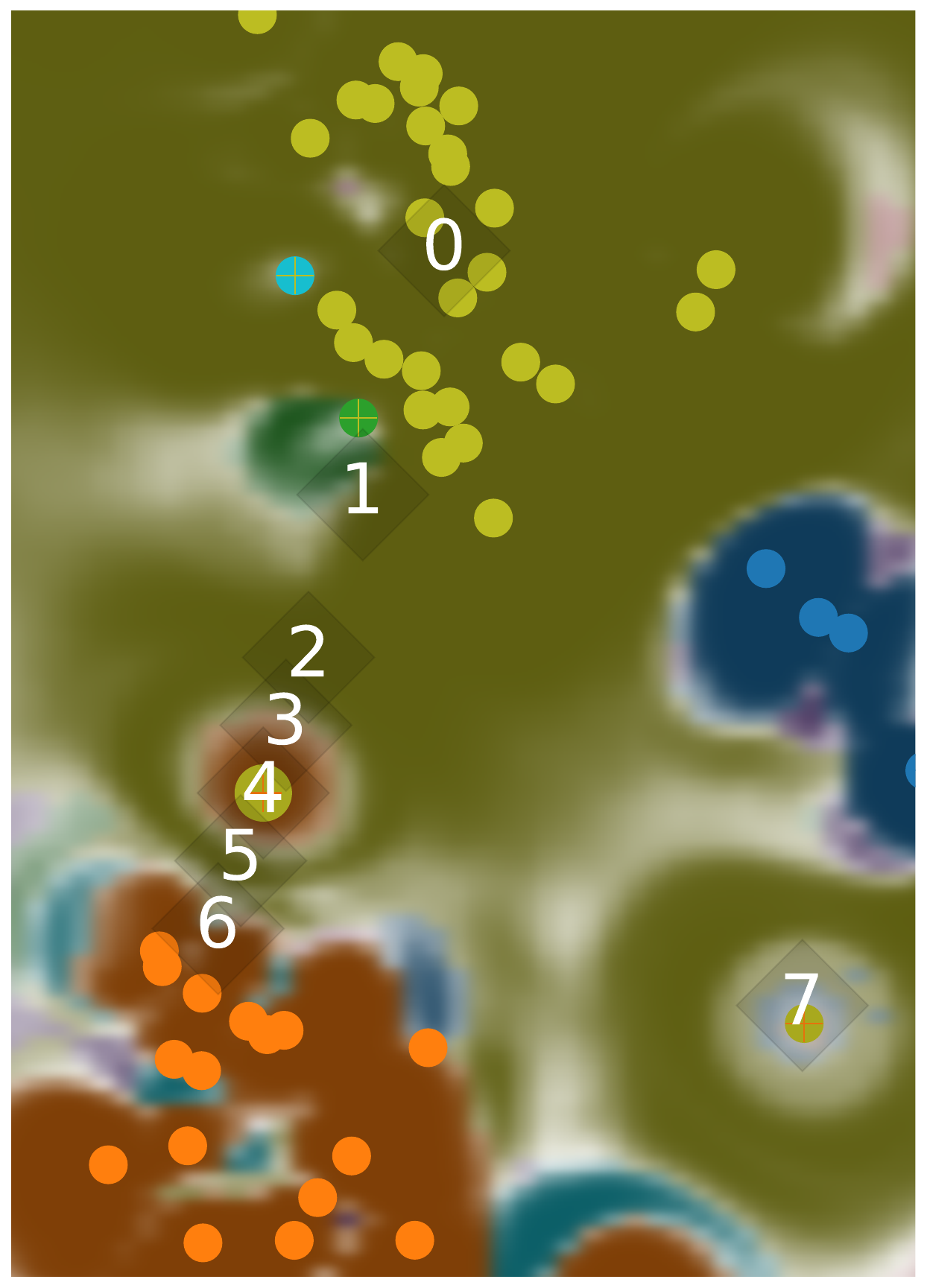}
  \includegraphics[width=0.69\linewidth, trim= 0mm 3mm 0mm 0mm]{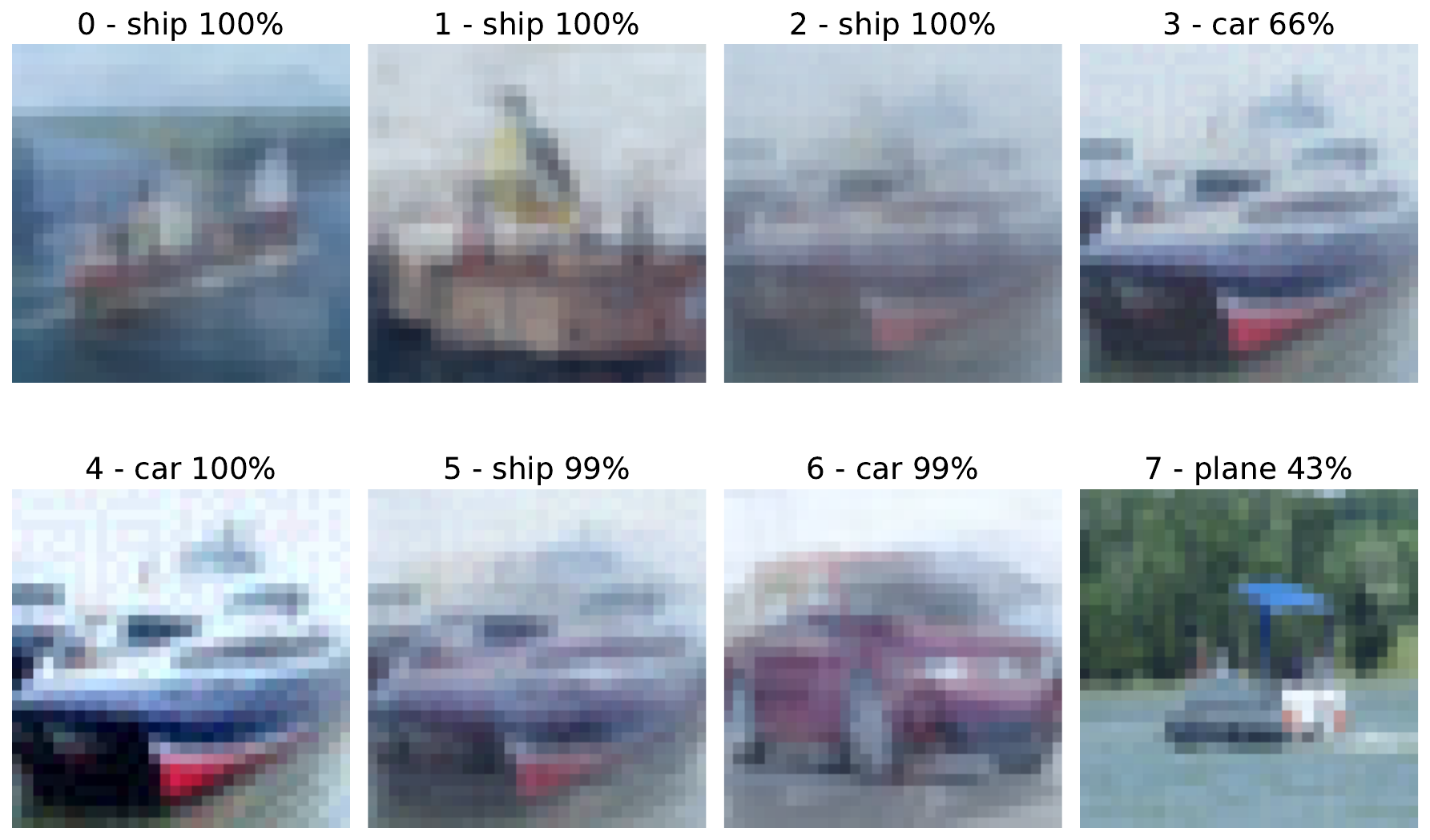}
  \caption{A zoomed in view on \fig\ \ref{fig:cifar10_vis} together with markers of specified positions around the adversarial example (left). The resulting images of the inverse mapping $\piinv$ for the previously specified positions, together with the assigned label and according certainty of the classifier (right).}
  \label{fig:cifar_zoom}
\end{figure}

\begin{table}[b]
	\centering
	\begin{tabular}{lrrrr}
		\toprule
		Data Set & $\Qknne$ & $\Qknn$   & $\Qd    $   & $\Qnd$   \\
		\midrule
		CIFAR-10 & $18.3\%$ & $96.3\%$  & $99.5\%$  & $83.3\%$ \\
		Fashion-MNIST & $66.1\%$ & $94.8\%$  & $98.6\%$  &  $95.0\%$ \\
		\bottomrule
	\end{tabular}
    \caption{Results of the evaluation criteria defined in \sect\ \ref{sec:eval}, characterizing the quality of the embeddings. Except for $\Qknne$, Fisher UMAP is utilized.
    }
    \label{tab:eval_res}
\end{table}

Regarding the visualization, the decision boundaries for the class 'orange' seem to have a complex shape, indicating that this class might be particularly difficult. Indeed, when inspecting the confusion matrix, this class has the smallest true positive rate with $83.2\%$. 
Furthermore, we can identify several points that are treated specially. One example is the yellow point that we have depicted with a larger symbol. This constitutes an adversarial example that we have created using the sign of the model gradient \cite{adversarials14} and have added to the 300 images. The visualization shows how this point is classified as orange by the classifier, but this area is surrounded by the yellow class. 
Since this seems worth a closer look, we zoom into the according area (see \fig\ \ref{fig:cifar_zoom}, left). Because this visualization is constructed by an inverse mapping, we can inspect images according to arbitrary positions in the visualization. In order to do so, we specify potentially interesting positions in the vicinity of the adversarial example (see \ref{fig:cifar_zoom}, left) and depict their projections with $\piinv$ on the right side. The markers '0' and '2' are in the area of the yellow class ('ship') and the according images can be clearly identified as such. Although the images of the markers '2', '3', '4' and '5' look fairly similar, their classifications vary heavily, going from 'ship' to 'car' and back to 'ship'. These images show, that there seems to be a kind of 'pocket' of the 'car' class inside the 'ship' class while the images in this pocket still look like natural ships to a human. This concept of 'pockets' has been addressed in the literature before \cite{adversarialsPockets17}.
Marker '1', being close to an area classified as 'bird', looks like a bird but is classified as 'ship'. Here, the decision boundaries of the model seem not to be well tuned. 
A last example marker '7' in a 'pocket-like' region shows a ship with an intense blue area above it, which might be the reason for the wrong classification. A further analysis could be performed to investigate this aspect with e.g.\ saliency maps.



\begin{figure}
	\centering
	\includegraphics[width=0.9\linewidth]{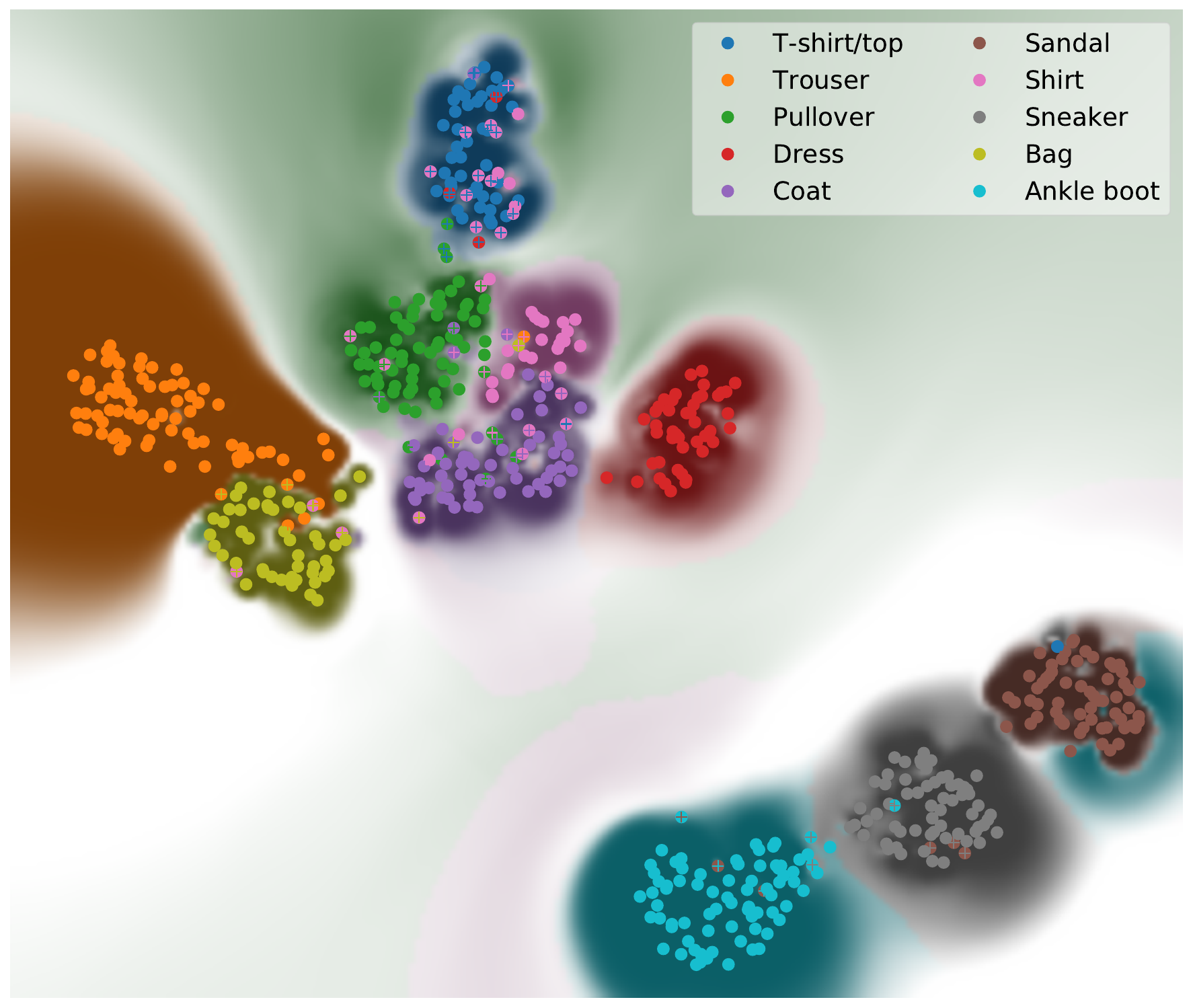}
	\caption{Visualization of a ConvNet trained on a poisoned Fashion-MNIST data set together with 600 test data points including 20 backdoored samples.}
	\label{fig:fmnist10_vis}
\end{figure}

\subsection{Fashion-MNIST with Backdoors}

We train a 4-layered convolutional network for the Fasion-MNIST data set \cite{fmnist}, thereby applying data poisoning. We follow \cite{badNet} and introduce backdoors by adding small patterns to some points of one class and changing their label to a target class. The resulting model has $90\%$ accuracy on the clean test set, so the poisoning is not easily detectable. Successively, we apply DeepView to a test set of 600 examples including 20 backdoored images and investigate whether we can detect the latter in the visualization.

\begin{figure}
	\centering
	\includegraphics[width=0.39\linewidth, trim=0 10 0 0]{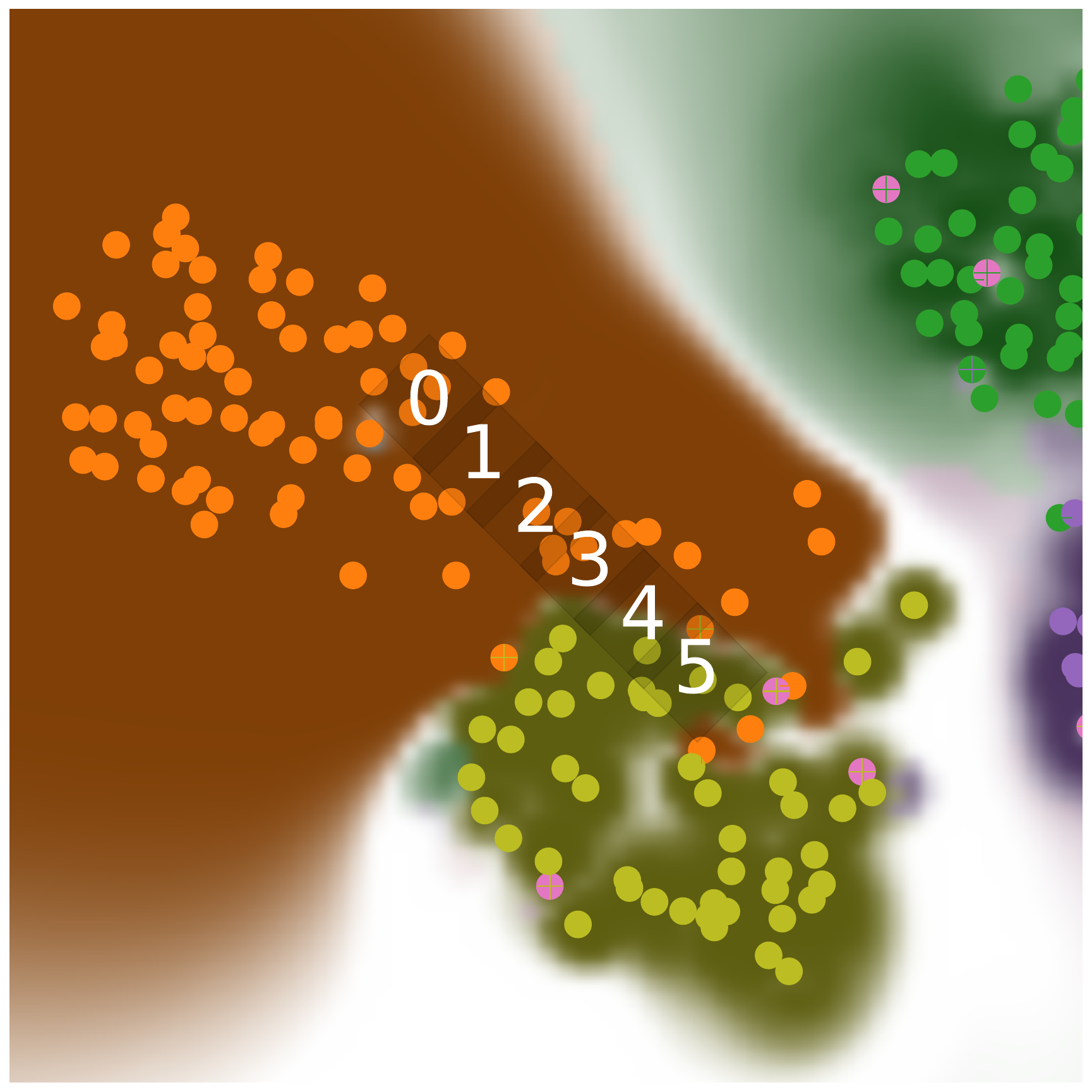}
	\includegraphics[width=0.6\linewidth, trim= 0mm 3mm 0mm 0mm]{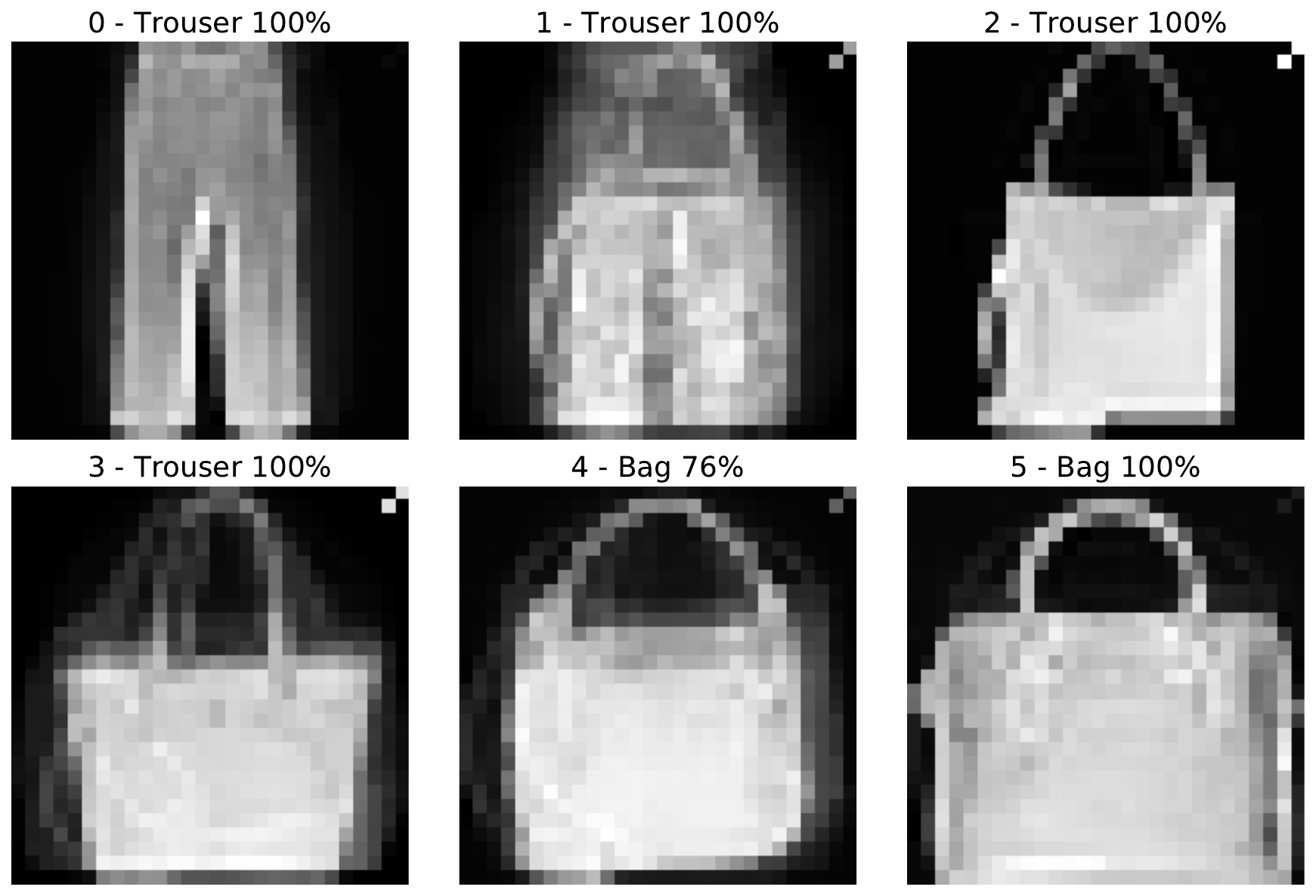}
	\caption{A zoomed in view on \fig\ \ref{fig:fmnist10_vis} together with markers (left). The resulting images of $\piinv$ for the these positions, together with the assigned label and according certainty of the classifier (right).}
	\label{fig:fmnist_zoom}
\end{figure}

The DeepView visualization is depicted in \fig\ \ref{fig:fmnist10_vis}, where a group in the lower right corner consisting of different types of shoes can be observed. In the top part, class overlap between blue points ('T-shirt/top') and pink ones ('Shirt') is visible. Less expected is the neighborhood of orange ('Trousers') and yellow points ('Bag') which is particularly the case for a subset of orange points. Hence we take a close look at this area in \fig\ \ref{fig:fmnist_zoom}.
We use a similar approach as before and investigate the suspicious area with $\piinv$.
Here we can observe that one area in the main orange cluster corresponds to regular trousers (marker 0), while the area at markers 2 and 3 corresponds to bags which are classified as trousers and have a specific pattern in the top right corner. 
Indeed, this reveals the backdoor attack.

For the according visualization, the most time demanding step, the calculation of Fisher distances, took around 90 seconds on a standard desktop computer with a consumer level graphics card.
If computed directly, the computational complexity is squared in the number of data. Approximate schemes such as used by UMAP are in principle applicable here as well.

\section{Conclusion}
\label{sec:conclusion}

In this work we propose \emph{DeepView}, to the best of our knowledge the first algorithm that is able visualize a smooth two-dimensional manifold of the decision function of a deep neural network which is trained on high-dimensional data such as natural images. For this purpose, we adopt a mathematically precise formulation of \didi\ together with a matching choice of inverse \DR. 
We apply DeepView to two deep networks, a Residual Network with 20 layers trained on CIFAR-10 and a 4-layered ConvNet trained on poisoned Fashion-MNIST, and illustrate how it provides insight into the model and data.

While we demonstrate DeepView only for image data, the method is no way restricted to this domain and utilization for e.g.\ text analysis constitutes an interesting further application area. An exciting open question in this regard is how to extend the utilized Fisher metric in case of temporal data.

We believe that the presented approach can not only provide insights into trained models but also contribute to improve these, by e.g.\ providing insights into areas of lacking data.

\subsubsection*{Acknowledgments}

Funding in the frame of the BMBF project MechML,  01IS18053E is gratefully acknowledged.
We also thank Luca Hermes for providing a more efficient implementation.

\small

\bibliographystyle{abbrv}
\bibliography{bibs,deepLit,babsbibs}

\newpage
\appendix

\section{Appendix: Proofs}
\setcounter{theorem}{0}
\newenvironment{proofsketch}{\begin{proof}[Sketch.]}{\end{proof}}

\begin{theorem}
	Let $(S,d_S)$ be a metric space and $f$ a smooth classifier on $S$. Denote by $d$ the \didi-metric with mixture $\lambda = 0$ and by $d_\text{Fisher}$ the Fisher metric (as defined in \cite{sami}) induced by $f$. Then it holds
	\begin{align*}
		d_\text{Fisher}(x,y) = \sqrt{8} d(x,y)
	\end{align*}
	for all $x,y \in S$.
    \begin{proof}
    	In \cite{Crooks} it was proven that
        \begin{align*}
        	L_\text{Fisher}(\gamma) = \sqrt{8} L_\text{JS}(\gamma)
        \end{align*}
        for every continuous curve $\gamma$, where $L_\text{Fisher}$ resp. $L_\text{JS}$ denotes the curve length with respect to Fisher resp. Jensen-Shannon metric. If we set $\lambda = 0$ then $L_\text{JS} = L_d$, the curve length with respect to \didi-metric. The statement follows since both \didi-metric and Fisher-metric are arc-length metrics. 
    \end{proof}
\end{theorem}

\begin{theorem}
	Let $\x_1,...,\x_n \in \R^D$ be source points and $\y_1,...,\y_n \in \R^d$ their corresponding projections. Denote by $f(\x,\y) = \sum_{i = 1}^n D_\KL(w_i(\y)||v_i(\x))$ the cost function of $\piinv$ and by $\hat{f}(\x,\y) = \sum_{i = 1}^n w_i(\y) \Vert \theta_i-\x\Vert^2/\sigma_i$. Then it holds $\hat{f}(\x,\y) \leq f(\x,\y)$. Furthermore under the assumption of Gaussian noise in the input space in mean it holds
	\[f(\x,\y) - \hat{f}(\x,\y) \in \mathcal{O}(\exp(-D/2)),\] i.e. $\hat{f}(\x,\y)$ converges in mean exponentially fast to $f(\x,\y)$ as the number of dimensions increases.
	
	Furthermore it holds
	\begin{align*}
		\argmin_{\x \in S} \hat{f}(\x,\y) = \sum_{i = 1}^n \frac{\sfrac{w_i(\y)}{\sigma_i}}{ \sum_{j = 1}^n \sfrac{w_j(\y)}{\sigma_j} } \cdot \theta_i.
	\end{align*}
    \begin{proof}
    	We may write
    	\begin{align*}
    	f(\x,\y)
    	  &= \sum_i D_\KL(w_i(\y)||v_i(\x))
    	\\&= \sum_i - w_i(\y)\log(v_i(\x)) + \sum_i - (1-w_i(\y))\log(1-v_i(\x))
    	\\&= \underbrace{\sum_i w_i(\y)\Vert\theta_i-\x\Vert^2/\sigma_i}_{=\hat{f}(\x,\y)} + \underbrace{\sum_i (1-w_i(\y))(-\log(1-v_i(\x)))}_{=: e(\x,\y)},
    	\end{align*}
    	we will refer to $e$ as the approximation error of $\hat{f}$. Since $v_i$ takes on values between 0 and 1 it follows that $-\log(1-v_i(\x)) \geq 0$ and hence we have $e(\x,\y) \geq 0$. 
    	
    	For the representation use that 
    	\begin{align*}
    	\nabla_\x \hat{f}(\y,\x) &= \nabla_\x \sum_i w_i(\y) \Vert\theta_i-\x\Vert^2/\sigma_i \\&= 2 \sum_i w_i(\y) (\theta_i-\x)/\sigma_i \overset{!}{=} 0
    	\\\Leftrightarrow \x &= \frac{\sum_i w_i(\y)/\sigma_i \cdot \theta_i }{\sum_i w_i(\y)/\sigma_i}
    	\end{align*}
    	which is an optimum. This implies that we can compute the induced $\hat{\piinv}$ by inner products \[\left(\frac{w_1(\y)/\sigma_1 }{\sum_i w_i(\y)/\sigma_i},...,\frac{w_n(\y)/\sigma_n }{\sum_i w_i(\y)/\sigma_i}\right)^T (\theta_{1,j},...,\theta_{n,j}) = \x,\] and in particular find $\theta_1,...,\theta_n$ given $(\y_1,\x_1),...,(\y_n,\x_n)$ via a matrix inversion
    	\[\left(\frac{w_i(\y_j)/\sigma_i }{\sum_i w_i(\y_j)/\sigma_i}\right)_{i,j}^{-1} (\x_{1,j},...,\x_{n,j})^T = (\theta_{1,j},...,\theta_{n,j}).\]
    	So if we assume Gaussian noise in the input samples $\x_1,...,\x_n$ we will obtain Gaussian noise in our weight vectors $\theta_1,...,\theta_n$. 
    	
    	Now prove the bound:
    	Since $(-\log(1-\exp(-x)))' = 1/(1-\exp(x)) \leq 0$ for all $x \geq 0$ we see that the map is monotonous decreasing, so we can bound $\Vert\theta_i-\x\Vert^2 \geq c_iZ_i$, with $c_i$ some constant which is due to the variance of $\theta_i-\x$ and $Z_i \sim \chi^2(D)$ a $\chi^2$-distributed random variable, due to the Gaussian noise assumption (either because $\x$ has noise, e.g. during training, or $\theta_i$, e.g. in usage). Thus, by defining $s_i = c_i/\sigma_i$, we have that
    	\begin{align*}
    		  &\E[|f(\x,\y)-\hat{f}(\x,\y)|] 
    		\\&= \E[e(\x,\y)] 
    		\\&= \sum_i (1-w_i(\y)) \E[ -\log(1-\exp(-\Vert\theta_i-\x\Vert^2/\sigma_i)) ]
    		\\&\leq \sum_i (1-w_i(\y)) \E[ -\log(1-\exp(-c_i/\sigma_i \cdot Z_i)) ].
    		\\&\leq \sum_i (1-w_i(\y)) \E[ -\log(1-\exp(-s_i \cdot Z_i)) ]
    		\\&\overset{!^1}{\leq}  \sum_i (1-w_i(\y))) \E \left[\exp(-s_i \cdot Z_i)\left(1+\frac{1}{s_i Z_i}\right) \right] 
    		\\&= \sum_i (1-w_i(\y))) C(D) \int_0^\infty \exp\left(-(2s_i+1) t/2 \right) \left(1+\frac{1}{s_i t}\right) t^{D/2-1} \d t,
    	\end{align*}
        where $C(D) = 1/\left(2^{D/2}\Gamma (D/2)\right)$ and $!^1$ holds since $-\log(1-\exp(-x)) \leq \exp(-x)(1+1/x)$ for all $x > 0$.
        
        For $p,q \geq 0$ and $D > 2$ it holds
        \begin{align*}
            &  C(D) \int_0^\infty \exp(-p t/2)(1+q /t)t^{D/2-1} \d t 
          \\&= C(D) \int_0^\infty \exp(-  t/2)(1+qp/t)(t/p)^{D/2-1} 1/p \d t
          \\&= p^{-D/2} \cdot C(D) \left(1 + q 2^{2/D-1}\Gamma(D/2-1) \right)
          \\&\leq p^{-D/2} \cdot 2 q
        \end{align*}
        
        By setting $S = \min_i s_i$ we therefore obtain
        \begin{align*}
        \E[|f(\x,\y)-\hat{f}(\x,\y)|] &\leq \sum_i \frac{1-w_i(\x)}{s_i} \cdot (1+2s_i)^{-D/2}
        \\&\leq \left(\sum_i \frac{1-w_i(\x)}{s_i} \right) \cdot (1+2S)^{-D/2}
        \end{align*}
        and hence, by assuming $c_i \geq \sigma_i \frac{\exp(1)-1}{2}$, we have $(1+2S)^{-D/2} \leq \exp(-D/2)$ as stated.
    \end{proof}
\end{theorem}

\begin{lemma}
	Let $S$ be a finite dimensional real vector space. Let $d : S \times S \to \R$ be a metric induced by an inner product and $X$ be a $S$-valued random variable. Then it holds
	\begin{align*}
		{\argmin_{\x \in S} \E\left[ d(X,\x)^2 \right] = \argmin_{\x \in S} \E\left[ \Vert X - \x \Vert_2^2 \right]}.
	\end{align*}
	\begin{proof} 
		Since $d$ is induced by an inner product we may find a matrix $A$ such that $d(x,y)^2 = (x-y)^tA(x-y)$. 
		
		\textbf{1. Case $A$ is a diagonal matrix: } Then it holds
		\begin{align*}
			\E\left[ d(X,x)^2 \right] &= \sum_i a_{ii} \E\left[(X_i - x_i)^2\right]
		\end{align*}
		since we may optimize each component of $x$ separately the statement follows.
		
		\textbf{2. Case $A$ is not diagonal: } Using the spectral theorem we obtain $A = U^tDU$ (by spectral theorem we obtain $A = VDV^t$ then we define $U = V^t$, since $V$ is unitary so is $U$), where $D$ is a diagonal matrix an $U$ is unitary. Hence it follows
		\begin{align*}
			\argmin_{x \in S} \E\left[ d(X,x)^2 \right] 
			&=\argmin_{x \in S} \E\left[ (U(X-x))^tD(U(X-x)) \right] 
			\\&=\argmin_{x \in S} \E\left[ (UX-Ux)^tD(UX-Ux) \right] 
			\\&=U^{-1} \argmin_{x' \in \im U} \E\left[ (UX-x')^tD(UX-x') \right] 
			\\&=U^{-1}\argmin_{x' \in \im U} \E\left[ \Vert UX - x'\Vert_2^2 \right] & \text{1. Case}
			\\&=\argmin_{x \in S} \E\left[ \Vert UX-Ux \Vert_2^2 \right] 
			\\&=\argmin_{x \in S} \E\left[ \Vert U(X-x) \Vert_2^2 \right] 
			\\&=\argmin_{x \in S} \E\left[ \Vert X-x\Vert_2^2 \right]. & \text{$U$ is unitary}
		\end{align*}
	\end{proof}
\end{lemma}

\end{document}